\pdfoutput=1
\documentclass[conference]{IEEEtran}
\IEEEoverridecommandlockouts
\usepackage{cite}
\usepackage{diagbox}
\usepackage{amsmath,amssymb,amsfonts}
\usepackage{algorithmic}
\usepackage[linesnumbered,ruled,vlined]{algorithm2e}
\usepackage{graphicx}
\usepackage{multirow}
\graphicspath{{./figures/}}
\usepackage{textcomp}
\usepackage{xcolor}
\def\BibTeX{{\rm B\kern-.05em{\sc i\kern-.025em b}\kern-.08em
		T\kern-.1667em\lower.7ex\hbox{E}\kern-.125emX}}
\begin{document}

\title{Improving the Performance of a NoC-based CNN Accelerator with Gather Support\\}

\author{\IEEEauthorblockN{Binayak Tiwari\IEEEauthorrefmark{4}, Mei Yang\IEEEauthorrefmark{4}, Xiaohang Wang\IEEEauthorrefmark{2}, Yingtao Jiang\IEEEauthorrefmark{4}, Venkatesan Muthukumar\IEEEauthorrefmark{4}}
	\IEEEauthorblockA{\IEEEauthorrefmark{4}\textit{Department of Electrical and Computer Engineering, University of Nevada}, Las Vegas, USA\\ \IEEEauthorrefmark{2}\textit{School of Software Engineering, South China University of Technology}, Guangzhou, China\\
		Email: \IEEEauthorrefmark{4}btiwari@unlv.nevada.edu, \IEEEauthorrefmark{4}\{mei.yang, yingtao.jiang, venkatesan.muthukumar\}@unlv.edu, \IEEEauthorrefmark{2}xiaohangwang@scut.edu.cn}
}

\maketitle

\begin{abstract}
	The increasing application of deep learning technology drives the need for an efficient parallel computing architecture for Convolutional Neural Networks (CNNs). A significant challenge faced when designing a many-core CNN accelerator is to handle the data movement between the processing elements. The CNN workload introduces many-to-one traffic in addition to one-to-one and one-to-many traffic. As the de-facto standard for on-chip communication, Network-on-Chip (NoC) can support various unicast and multicast traffic. For many-to-one traffic, repetitive unicast is employed which is not an efficient way. In this paper, we propose to use the gather packet on mesh-based NoCs employing output stationary systolic array in support of many-to-one traffic. The gather packet will collect the data from the intermediate nodes eventually leading to the destination efficiently. This method is evaluated using the traffic traces generated from the convolution layer of AlexNet and VGG-16 with improvement in the latency and power than the repetitive unicast method.
\end{abstract}


\section{Introduction}
Deep learning technology has made tremendous improvements in solving problems like pattern recognition, object detection, forecasting, etc. A popular class of deep learning architecture used in computer vision and image processing field is a Convolutional Neural Networks (CNNs) which may contain a large number of layers and neurons. CNN models such as AlexNet \cite{alexnet} and VGG-16 \cite{vgg16} each consists of millions of parameters and tens of layers. CNN training/inference process involves a large number of complex vector and matrix computations with a high degree of data parallelism. The parallel computation and communication patterns involved in these CNNs impose the need for efficient parallel computing architectures \cite{didianno}.

Network-on-Chip (NoC) has emerged as the de-facto standard for on-chip communication in multi/many-core systems \cite{dally}. The need for NoC in a CNN accelerator is ever increasing and becoming important because of the inherent nature of CNN computation and the on-chip traffic involved. Another important feature of NoC is scalability where nodes or processing elements (PE) can be added with minimal changes. Modular design property helps in gating and thus saving the power by turning off the unused modules. This also provides flexibility in running different kinds of workloads on the same NoC-based system \cite{nocbook}.

Different from conventional parallel workloads like PARSEC \cite{parsec}, CNNs involve a significant amount of many-to-one traffic in addition to one-to-one and one-to-many type of traffic. An example of many-to-one traffic is in the convolution layers in a CNN model when the partial sums are returned from the processing cores to the memory. Besides, the communication pattern on CNNs is mostly repetitive because of the large number of computations being executed on a resource limited PEs. Data communication in a CNN accelerator consumes around 30\% of the total energy and increases with the system scaling \cite{didianno}.

In this paper, we propose to use the gather packet to support many-to-one type of traffic for a mesh-based NoC which has not yet been addressed. Based on the output stationary systolic array, the gather packet will collect the data from the intermediate nodes eventually leading to the destination efficiently. This method is evaluated using the traffic traces generated from the convolution layer of AlexNet and VGG-16 with improvement in the latency and power than the repetitive unicast method.

The rest of the paper is organized as follows. Section \ref{background} provides relevant background, literature review and the motivation. Section \ref{proposed} describes the proposed routing algorithm and packet format to support many-to-one traffic. Section \ref{router_arch} describes the proposed changes needed in the router microarchitecture. Section \ref{performance} discusses the results obtained and Section \ref{conclusion} concludes the paper.

\section{Background and Motivation} \label{background}
A CNN may consist of multiple alternate layers (like convolution and maxpooling) which perform the task of extracting feature maps followed by the final fully connected layers which perform the task of classification. Input to a convolution layer can be a raw image or input feature map which is essentially the output of the previous convolution layer. This input is convolved with a set of filters to produce an output feature map. With the increasing layers in recent CNN models, the number of convolution layers is also increasing with convolution operations accounting for over 90\% of the overall operations \cite{eyeriss}. Several CNN accelerators \cite{didianno}, \cite{eyeriss}, \cite{rethinking}, \cite{fpgaaccelerator} have been proposed recently. Traffic on a CNN accelerator can be classified as one-to-one or unicast traffic, one-to-many or scatter kind traffic like the traffic from memory or a buffer to the processing elements (PEs), and many-to-one or gather kind traffic such as the traffic from PEs to memory or a buffer. The scatter kind traffic can be supported using multicast packets. The gather kind traffic may occur when the current PE operation is finished and the obtained partial sums need to be stored for the next set of PE operation or in general storing the PE result for future need.

Various methods have been proposed in the literature to support the CNN traffic with a multi/many-core NoC based system. Authors in \cite{neunoc} proposed an NN-aware mapping to reduce the distance of communication and developed a multicast packet to reduce the redundant traffic. Authors in \cite{mappingsolution} also proposed a mapping solution to map different layers to a network based on the communication and computation constraint. Authors in \cite{flowmapping} proposed a traffic distribution approach based on the memory access mechanism using a distributer node to reduce the latency and energy consumption in a mesh network. Authors in \cite{reconfigurable} proposed a reconfigurable NoC which is capable of handling the multicast traffic in the NN accelerators, they proposed a bus-based reconfigurable NoC which adapts topology to the network traffic pattern. Authors in \cite{rethinking} proposed an NoC designed for the traffic flow in the NN using an array of microswitches that handle gather and scatter traffic by creating a sub tree from the buffer to the PE array.

Mesh is the most commonly used NoC topology which is scalable and can support a high number of PEs. A variety of deadlock-free unicast and multicast routing algorithms have been proposed for mesh-based NoCs. Our study is focused on providing efficient routing support for many-to-one traffic on mesh-based NoCs. Systolic architecture \cite{systolic} is an efficient way of implementing the convolution operation in hardware. In \cite{eyeriss}, various systolic CNN architectures are described which can be applied to mesh-based NoCs. In Weight Stationary (WS) systolic array, each PE gets the weight from the memory or buffer and keeps the weight until all the computation related to the weight is finished. While in Output Stationary (OS) systolic array, weight and input are streamed into PE every cycle to obtain the partial sums and No Local Reuse (NLR). These systolic architectures \cite{systolic} reuse the data maintaining the uniform dataflow within each layer. In this paper, a mesh topology with the dataflow model of output stationary is considered.

\begin{figure}
	\centering
	\includegraphics[width=1\linewidth]{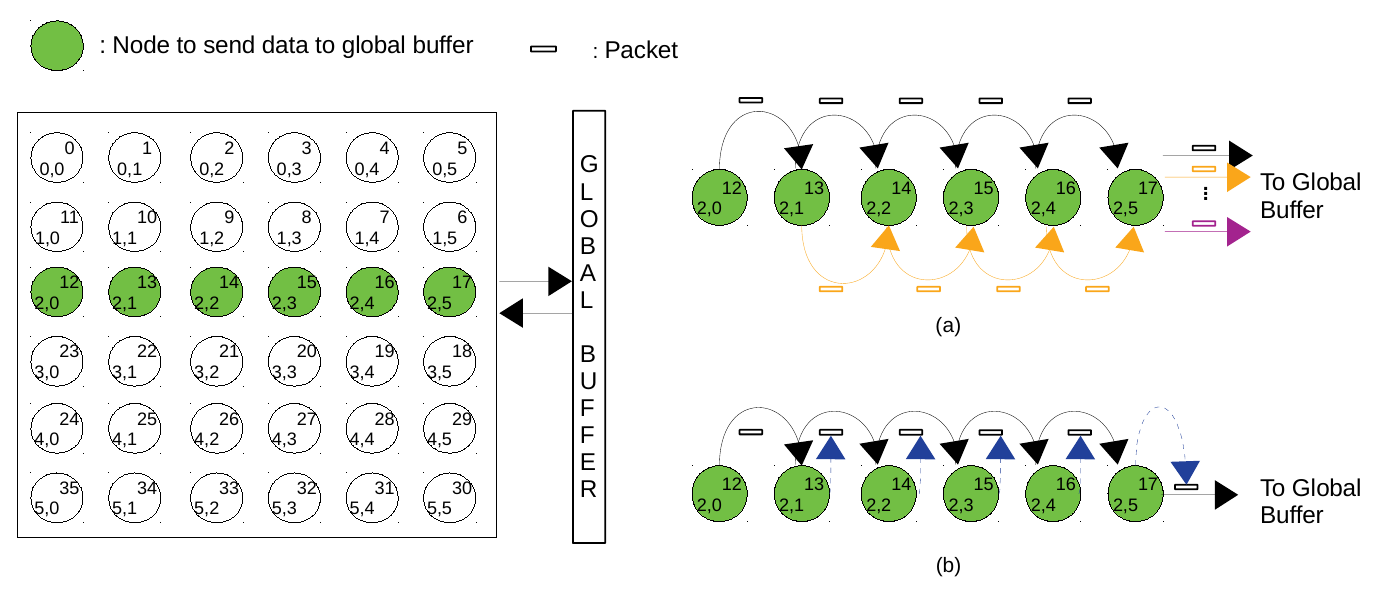}
	\caption{6x6 mesh example (a) without gather support (b) with gather support}
	\label{fig-motivation}
\end{figure}

Fig. \ref{fig-motivation} shows a 6x6 mesh with six PEs ready to send data to the global buffer, Fig. \ref{fig-motivation}(a) shows the operation without the gather support where each PE sends a unicast packet to the global buffer which requires 15 hops to get the data to the buffer. Is it possible to collect the data to be sent to the buffer into one gather packet? Fig. \ref{fig-motivation}(b) illustrates the gather packet initiated from node 12. As it progresses along the route, the gather packet will enclose the data payload from each intermediate node. The hop count for the gather packet is only 5. It is clear that the gather packet is able to deliver the data to the global buffer with less resource utilization by removing the redundant use of the resources compared to the repetitive unicast method. The details of the proposed method are described in Section \ref{proposed}.

\section{Proposed Method} \label{proposed}
In this section, the NoC architecture and data flow on a systolic array will be introduced followed by the description of the gather packet support routing scheme and analysis of the performance gain. 
\subsection{Data Flow Model}
\begin{figure}
	\centering
	\includegraphics[width=1\linewidth]{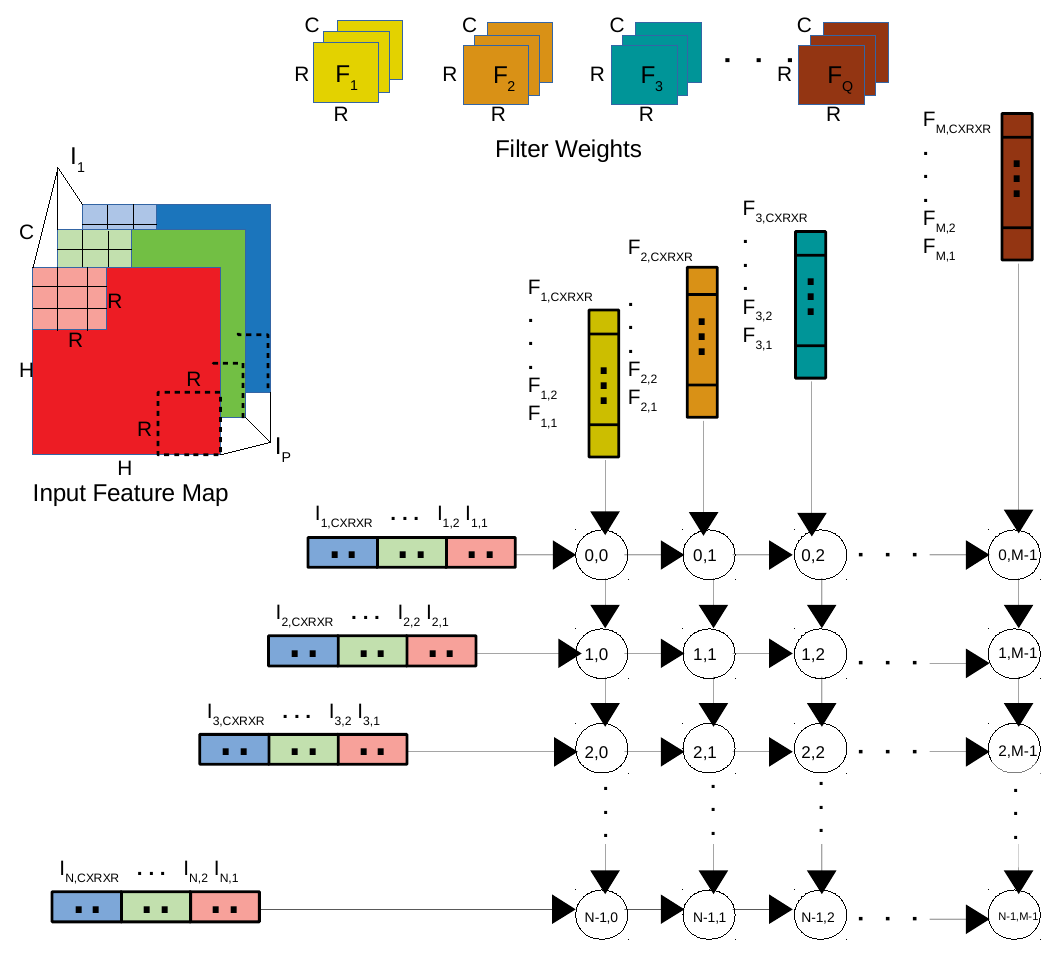}
	\caption{Dataflow for output stationary (OS)}
	\label{fig-os}
\end{figure}

The mesh-based NoC adopting the Output Stationary (OS) \cite{rethinking} \cite{eyeriss} systolic architecture is shown in Fig. \ref{fig-os}. Each PE can perform the simple multiply and accumulate (MAC) operation. The input data and filter weights are fed from the left and up edges to the left-most column PEs and up-most row PEs, respectively. The output data will be stored to the global buffer located on the right side of the NoC which will be accessed by the subsequent layers. These PEs will propagate the data to the PEs on the same column and row until all the PEs receive the data. One of the advantages of systolic architecture is that the whole operation is repeated in the pipelined manner making it easier to perform input and output operation. In each pipeline stage, each PE stores the data received from its left and up neighboring nodes, performs the MAC operation, and forwards the same data to the right and down neighbor nodes in parallel. This way helps in reducing a significant amount of memory operations.

Without the loss of generality, assume that there are $N \times M$ PEs in the mesh-based NoC. The $P$ inputs or feature maps of size $H \cdot H$ with $C$ channels is streamed from the left edge of the NoC and $Q$ filters each of size $R \cdot R$ with $C$ channel are streamed from the top edge. Each PE will receive a total of $C \cdot R \cdot R$ input and weight data then perform $C \cdot R \cdot R$ MAC operations. On this setting $PE_{0,0}$ will be the first to
finish the operation and then $PE_{1,0}$, $PE_{0,1}$ are the second in line and so on. When all $P$ inputs are convolved with $Q$ filters the convolution operation is complete. Noticeably $P \cdot Q$ is not a trivial number and the convolution will be completed by $N \times M $ PEs in multiple rounds. Thus, once the $C \cdot R \cdot R $ data is streamed, partial convolution (PC) output as shown in (\ref{eq1}) is sent to the global buffer before the next set of data is streamed.

\begin{equation} \label{eq1}
\underset{i \in \{1...P\};k \in \{1...Q\}}{PC_{i,k}} = \sum_{j=1}^{C \cdot R \cdot R} (I_{i,j} \cdot F_{k,j} )
\end{equation}

\subsection{Gather Packet Support Routing Scheme}
\begin{figure}
	\centering
	\includegraphics[width=1\linewidth]{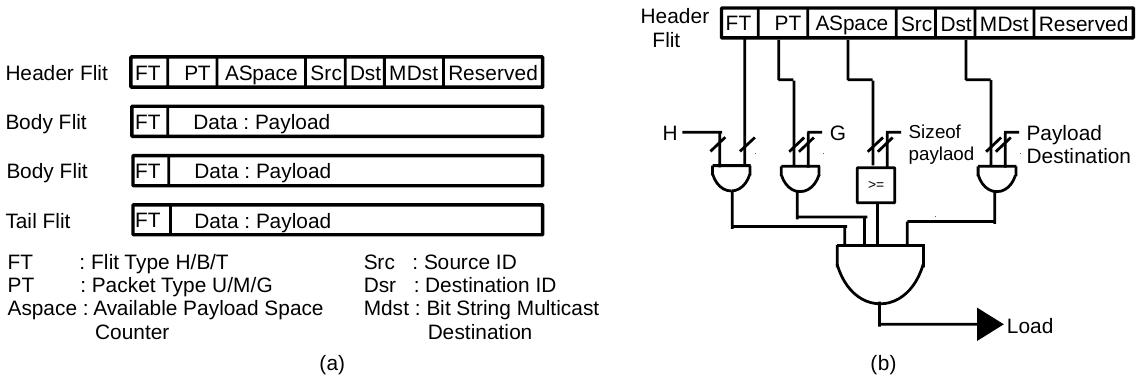}
	\caption{(a) Packet format (b) Generating gather load signal}
	\label{fig-packet}
\end{figure}

The process of moving the data from the PE to memory is gather traffic or many-to-one traffic. In this paper, we propose a gather traffic support routing scheme. The PE which has finished the operation first will initiate the gather packet towards the memory or a global buffer and on its way, it will collect all the available results from the intermediate nodes on the same row until its capacity is reached. Fig. \ref{fig-packet}(a) shows the packet format for the proposed gather support. \textit{FT} is the flit type that includes the head, body, and tail identifier. \textit{PT} is the packet type that includes unicast, multicast, and gather type identifier. \textit{ASpace} in the header flit is used to identify available space in a packet for a node to upload its payload. \textit{Src} and \textit{Dst} are the identifiers for source and destination node and finally, \textit{MDst} is the bit string multicast destination representation. Both body and tail flits include the payload field.

\begin{algorithm}
	\DontPrintSemicolon
	\SetAlgoLined
	\SetNoFillComment
	
	%\SetSideCommentLeft
	\SetKwInOut{Input}{Input}
	\SetKwInOut{Output}{Output}
	\Input{Arriving flit ($F$), Gather Payload ($P$) }
	\Output{Updated flit ($F$) or initiate a gather packet}

	\If{(($F.FT$ = $H$) and ($F.ASpace$ $>=$ $sizeof(P)$) and ($F.PT$ = $G$))}
	{	\tcp*[h]{generate a load signal}  
		
		\lIf{($F.Dst$ = $P.Dst$)} 
		{Load $\leftarrow$ 1} 
		\tcp*[h]{update F.ASpace before switch traversal}
		
		\lIf{(Load = 1)}{
			$F.ASpace \leftarrow$ $F.ASpace$ - $sizeof(P)$
		}
	}
	\lIf{(($F.FT$ = $B\ or\ F.FT$ = $T$) and (Load = 1))}
	{
		$F.Data \leftarrow$ $P.Data$
	}
	\lElse{$Can\ initiate\ a\ gather\ packet\ after\ \delta\ clock\ cycles$}
	\lIf{($F.FT$ = $T$)}
	{
		Load $\leftarrow$ 0
	}
	\vspace{-0.1cm}
	\caption{Flow for the Gather Support}
	\label{algorithm-flow}
\end{algorithm}

Algorithm \ref{algorithm-flow} shows the implementation of the gather support. Assume that the load signal is initialized to 0.  Upon the reception of a header flit in the input buffer, the router will identify if the packet is of gather type, verifies if there is space available for the payload, checks the destination address, and generates the load signal. Generally, the size of a gather payload is less than or equal to the flit size. Ideally, we want to upload the payload in one single flit. Fig. \ref{fig-packet}(b) shows the logic for generating a gather load signal. This signal is used to decrement the available space counter (\textit{ASpace}) for the payload which will be uploaded in the subsequent body or tail flit. 

Different conditions can arise which has to be taken care of for the smooth operation of the gather traffic. If \textit{ASpace} is zero, i.e, there is no space for a router to upload payload, the router will initiate its gather packet. This condition is covered in Algorithm \ref{algorithm-flow}. To avoid the flooding of the gather packets, a new gather packet will be generated after a $delta(\delta)$ cycle interval, which can be set depending on the router pipeline delay to reach the neighboring node and this can be configured for each router. This will allow sufficient time for the node to wait for the incoming gather packet from the neighboring node. 
\begin{figure}
	\centering
	\includegraphics[width=1\linewidth]{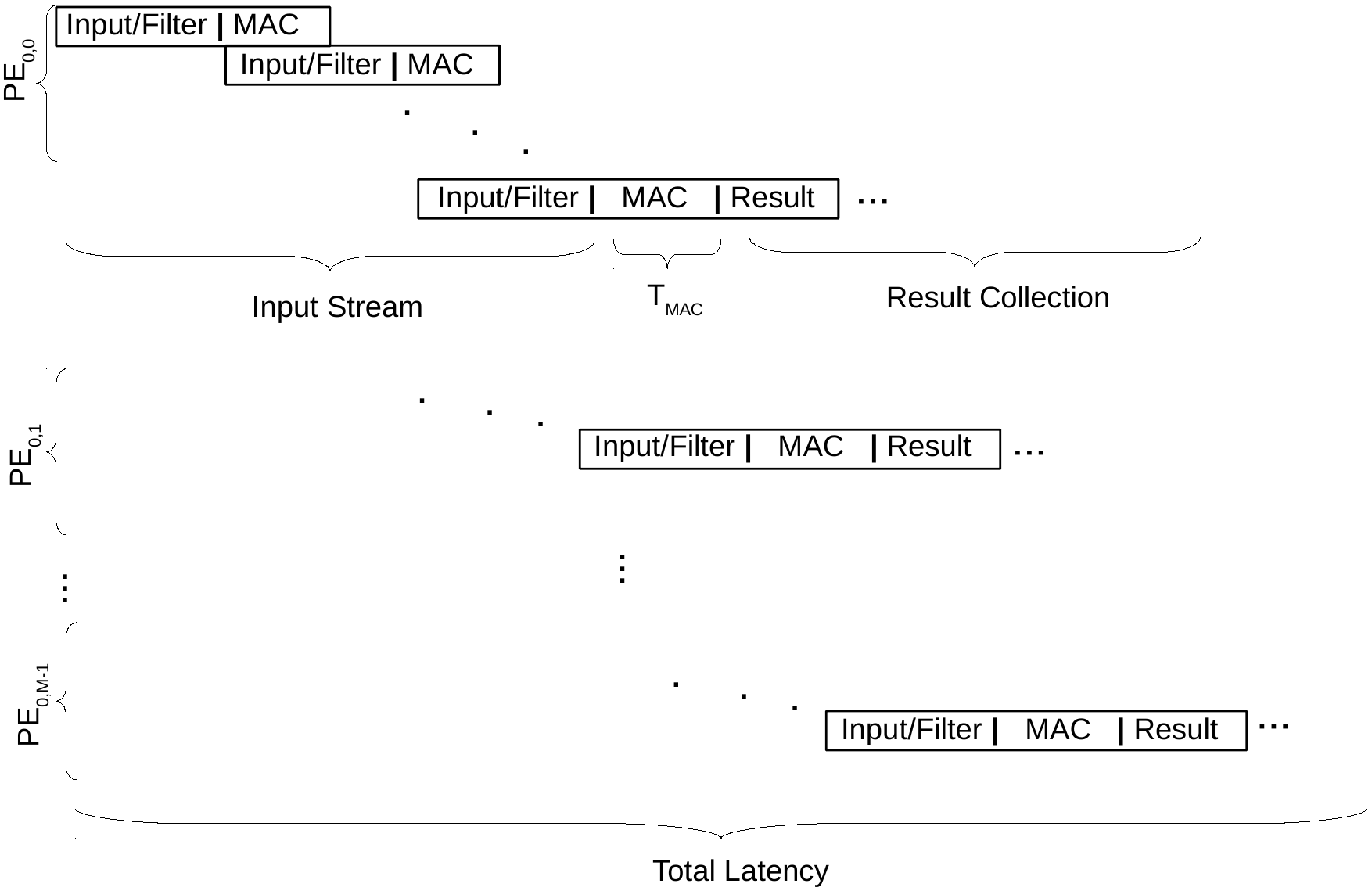}
	\caption{Pipelined operation of convolution on a row of PEs}
	\label{fig-latency_chart}
\end{figure}      
\subsection{Analysis of Performance Improvement}
The advantage of the proposed method is the savings in network latency and power consumption that are obtained using the gather packet to collect data that were otherwise going to be sent using many-to-one traffic. The gather packet reduces the hop count needed to deliver the data to the memory by allowing the data from the intermediate nodes to piggyback its payload into the incoming gather packet. 

The latency to finish one round of convolution is attributed to the input streaming time, the MAC computation time, and the result collection time. Assume that wormhole switching is used and the operations of input streaming/MAC/result collection are performed in fully pipelined fashion as shown in Fig. \ref{fig-latency_chart}. Equations (\ref{eq2}) and (\ref{eq3}) estimate the latency in clock cycles using the repetitive unicast (RU) method and the proposed gather method (G) respectively. Equation (\ref{eq4}) shows the expected performance improvement using the proposed gather method over the repetitive unicast.

In these equations, $C \cdot R \cdot R$ represents the time to stream the inputs to the PE, $T_{MAC}$ represents the computation time for the MAC operation, $\kappa$ represents the number of pipeline stages at each router (each stage occuping one cycle),  $\frac{P }{N} \cdot \frac{Q}{M}$ represents the multiple rounds for all the $P$ inputs and $Q$ filters. Assume that each unicast packet size is $L$, each gather packet size is $L'$, and the flit size is $W$. The gather packet is initiated from the leftmost node of each row.

\textit{$Latency_{_{RU}}$ =} 
\begin{equation} \label{eq5}
\resizebox{1\hsize}{!}{$
\Big( C \cdot R \cdot R + T_{MAC}  + M \cdot \kappa +  \Big \lceil \frac{L}{W} \Big \rceil -1  + (M-1) \cdot \Big \lceil \frac{L}{W} \Big \rceil +  \Delta_R \Big)\frac{P}{N} \cdot \frac{Q}{M}
\nonumber$}
\end{equation}

\textit{$=$} 
\begin{equation} \label{eq2}
\resizebox{1\hsize}{!}{$
	\Big( C \cdot R \cdot R + T_{MAC}  + M \Big(  \kappa +  \Big \lceil \frac{L}{W} \Big \rceil \Big )  - 1 +  \Delta_R \Big)\frac{P}{N} \cdot \frac{Q}{M}
	$}
\end{equation}
where $M \cdot \kappa$ represents the latency for the header flit from $PE_{0,0}$ to reach to the global buffer, $\Big \lceil \frac{L}{W} \Big \rceil -1$ represents the remaining flits from  $PE_{0,0}$ to reach the global buffer, $(M-1) \cdot \Big \lceil \frac{L}{W} \Big \rceil $ represents the latency for the remaining packets to reach to the global buffer, and $\Delta_R$ is the latency added due to the congestion. 
 \\

\textit{$Latency_{_{G}}$ =} 
\begin{equation} \label{eq3}
\resizebox{1\hsize}{!}{$
\Big(C\cdot R \cdot R+ T_{MAC}+\sum_{i = 0}^{\big \lceil \frac{M}{\eta}\big \rceil - 1}  \big(( M - i \cdot \eta) \cdot \kappa + \big \lceil \frac{L'}{W} \big \rceil -1 +t_\delta+  \Delta_G \big)\Big)\frac{P}{N} \cdot \frac{Q}{M}  $}
\end{equation}
where $\eta$ is the number of payloads that can be collected by one gather packet, $\big \lceil \frac{M}{\eta}\big \rceil $ represents the number of gather packet, $(M - i \cdot \eta) \cdot \kappa$ represents the latency for the header flit in the gather packet, $\big \lceil\frac{L'}{W} \big \rceil-1$ represents the latency for the rest of the flit in the gather packet, $t_\delta \in \{0,1,..,\delta\}$ is the latency added due to the $delta(\delta)$ cycle which will vary depending on the availability of gather packet when the gather payload is ready, and $\Delta_G$ is the latency added due to the congestion.
 \\

\textit{$Improvement$ =} 
\begin{equation} \label{eq4}
\resizebox{1\hsize}{!}{ $
\frac{\Big( M \big(  \kappa +  \big \lceil \frac{L}{W} \big \rceil \big ) -1 +  \Delta_R \Big) - \Big(  \sum_{i = 0}^{\big \lceil \frac{M}{\eta} \big \rceil - 1}  \big(( M - i \cdot \eta) \cdot \kappa + \big \lceil \frac{L'}{W} \big \rceil -1 +t_\delta+  \Delta_G \big)  \Big)}{C\cdot R \cdot R+ T_{MAC}+\sum_{i = 0}^{\big \lceil \frac{M}{\eta}\big \rceil - 1}  \big(( M - i \cdot \eta) \cdot \kappa + \big \lceil \frac{L'}{W} \big \rceil -1 +t_\delta+  \Delta_G \big)  }
$}
\end{equation}

\section{Rouer Architecture} \label{router_arch}

\begin{figure}
	\centering
	\includegraphics[width=0.9\linewidth]{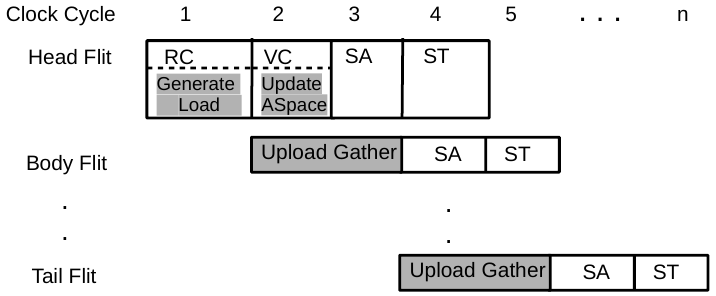}
	\caption{Modified router pipeline}
	\label{fig-pipeline}
\end{figure}

The gather payload can be uploaded into the incoming body and tail flits of the gather packet without letting the gather packet exit the router but fully utilize the router pipeline. Fig. \ref{fig-pipeline} shows how the four-stage router pipeline is used where header flit undergoes all the stages i.e, Route Computation (RC), Virtual Channel Allocation (VC), Switch Allocation (SA) and Switch Traversal (ST). However, body and tail flit does not perform any operation on the RC and VC stages of the router pipeline. These two pipeline stages are used by the router to upload the gather payload to the gather packet. 

Upon the arrival of a header filt, the load signal is generated during the RC stage and ASpace is updated in the VC stage from the information in the input buffer as shown in Fig. \ref{fig-packet}(b). On the subsequent arrival of body and tail flit, the RC and VC stages in the router pipeline are used to update the flit with the payload that is ready to be uploaded. This way, there is no impact on the router performance as the router pipeline is kept the same. 

\begin{figure}
	\centering
	\includegraphics[width=0.9\linewidth]{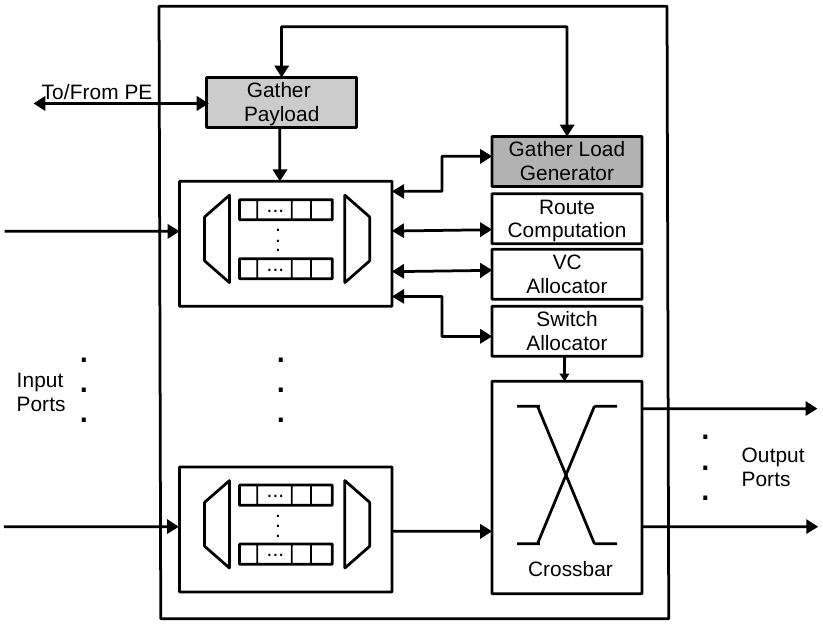}
	\caption{Router microarchitecture}
	\label{fig-microarchitecure}
\end{figure}

The proposed router microarchitecture is shown in Fig. \ref{fig-microarchitecure}. The Gather Load Generator block is responsible for generating the load signal (as shown in Fig. \ref{fig-packet}(b)) and updating ASpace in the header flit. The Gather Payload will hold the gather payload data obtained from the PE, sends the ack/nack back to the PE based on the status of payload upload. Failure to receive ack and on the expiration of the $delta(\delta)$ cycle, PE will initiate its gather packet.

\section{Performance Evaluation} \label{performance}
In this section, the proposed gather packet method is evaluated and compared with the repetitive unicast method using the convolution layers in AlexNet \cite{alexnet} and VGG-16 \cite{vgg16}.

\subsection{Simulation Settings}
A cycle accurate C++ based NoC simulator \cite{popnet} is used to simulate the proposed method implemented with the OS systolic array \cite{eyeriss} on mesh-based NoCs as shown in Fig. \ref{fig-os}. Table \ref{table-network} lists the NoC settings. Pytorch \cite{pytorch} framework is used to generate the parameters for AlexNet \cite{alexnet} and VGG-16 \cite{vgg16} shown in Table \ref{table-conv}. These parameters are used to generate the traces for the experiment. Orion 3.0 \cite{orion} is used to estimate the dynamic power consumption. $Delta (\delta)$ is set to the $5\ clock\ cycles$ to ensure at least the head flit will arrive at the neighboring node. Assume that the global buffer is connected to the right edge of the mesh. As shown in Fig. \ref{fig-os}, the input feature maps and filter weights are streamed from the input and weight buffers from the left and top side of the network respectively. A row- based gather is simulated where the gather packet is initiated from the PEs in the leftmost column to the rightmost PEs and then to the global buffer.

\begin{table}[]
	\caption{Network Configuration}
	\begin{center}
		\begin{tabular}{|l|l|}
			\hline
			Topology         & 8x8 Mesh, 16x16 Mesh                                                                    \\ \hline
			Virtual Channels & 4                                                                                       \\ \hline
			Router Pipeline Stage     & 5                                                                                 \\ \hline
			Buffer Depth     & 4 flits                                                                                 \\ \hline
			Packet Size      & \begin{tabular}[c]{@{}l@{}}Gather : 4 flits/packet\\ Other: 2 flits/packet\end{tabular} \\ \hline
			Flit Size        & 98 bits/flit                                                                            \\ \hline
			Gather Payload   & 32 bits                                                                                 \\ 
			\hline
			$T_{MAC}$   & 5 Clock Cycles                                                                                 \\ 
			\hline
		\end{tabular}
	\end{center}
	\label{table-network}
\end{table}

\subsection{Result}
\begin{figure}
	\centering
	\includegraphics[width=0.90\linewidth]{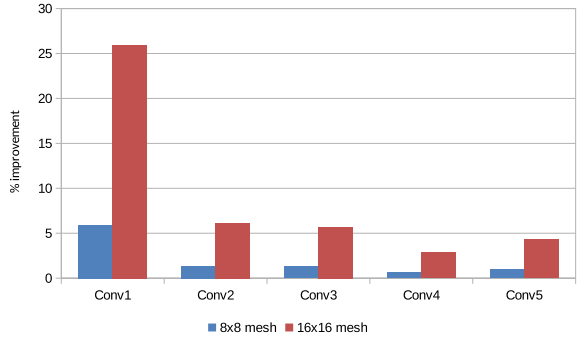}
	\caption{Improvement in total latency for AlexNet \cite{alexnet} on $8 \times 8$ and $16 \times 16$ mesh }
	\label{fig-latency-alexnet}
\end{figure}

Fig. \ref{fig-latency-alexnet} shows the performance improvement in the total latency (including the input streaming time, MAC time, and result collection time) of all five convolution layers in AlexNet \cite{alexnet} using the proposed gather method over the repetitive unicast method. Table \ref{table-est} shows that the changing trend of both results is consistent. For the estimated result in Table \ref{table-est},the parameters $\Delta_G$, $\Delta_R$, and $t_{\delta}$ are all set to 0, which reflects the ideal case with no congestion and $\delta$ delay. Noticeably, the estimated result represents the least improvement that can be achieved. Nevertheless, in simulations, there exists congestion and delta delays, i.e., $\Delta_R$, $\Delta_G$, and $t_\delta$ are not 0. That's why the simulated improvement is higher than the estimated result from (\ref{eq4}).

\begin{table}[]
	\caption{Estimated vs Simulated Performance Improvement in Total Latency for Alexnet \cite{alexnet} in 8x8 mesh}
	\begin{center}
		\begin{tabular}{|c|c|c|c|c|c|}
			\cline{1-6}
			\diagbox[width=0em]{Result}{Layers}& Conv1 & Conv2 & Conv3 & Conv4 & Conv5 \\ \hline
			\multicolumn{1}{|c|}{Estimated} & 2.92  & 0.73  & 0.68  & 0.34  & 0.51  \\ \hline
			\multicolumn{1}{|c|}{Simulated} & 5.93  & 1.37  & 1.27  & 0.63  & 0.95  \\ \hline
		\end{tabular}
	\end{center}
	\label{table-est}
\end{table}

\begin{figure}
	\centering
	\includegraphics[width=0.90\linewidth]{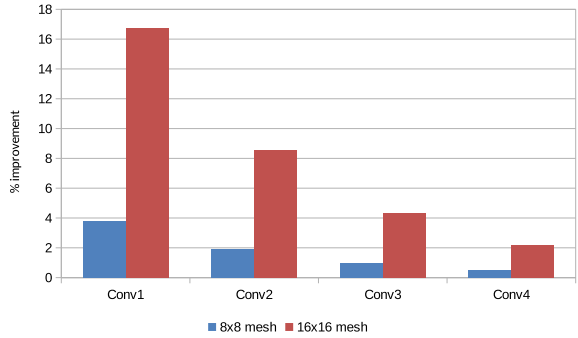}
	\caption{Improvement in total latency for VGG-16 \cite{vgg16} on $8 \times 8$ and $16 \times 16$ mesh }
	\label{fig-latency-vgg16}
\end{figure}

It's clear that $16 \times 16$ mesh offers better improvement than $8 \times 8$ mesh as the saving in hop count achieved by the gather method in $16 \times 16$ mesh is more significant than that in $8 \times 8$ mesh. In addition, the congestion will start to degrade the performance of the repetitive unicast with the increase in the network size.  The first convolution layer (Conv1) shows the highest improvement in both meshes because it has the smallest $C \cdot R \cdot R$ value (refer to Table \ref{table-conv} and (\ref{eq4})). Similarly, Fig. \ref{fig-latency-vgg16} shows the performance improvement in the total network latency for the selected 4 out of 13 convolution layers 2, 4, 6, 13 with different parameters in VGG-16 \cite{vgg16}. Similarly, Conv1 (for layer 2) which has the smallest $C\cdot R \cdot R$ value has the best improvement compared with other convoluntional layers.

\begin{figure}
	\centering
	\includegraphics[width=0.90\linewidth]{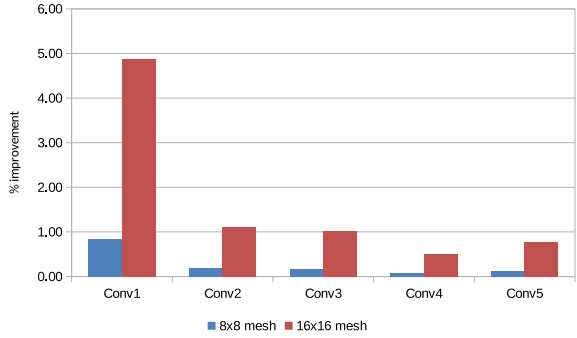}
	\caption{Improvement in power for AlexNet \cite{alexnet} on $8 \times 8$ and $16 \times 16$ mesh }
	\label{fig-power-alexnet}
\end{figure}

\begin{figure}
	\centering
	\includegraphics[width=0.90\linewidth]{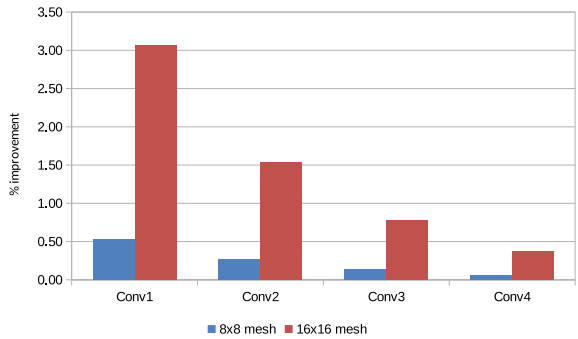}
	\caption{Improvement in power for VGG-16 \cite{vgg16} on $8 \times 8$ and $16 \times 16$ mesh }
	\label{fig-power-vgg16}
\end{figure}

Fig. \ref{fig-power-alexnet} shows the performance improvement in the total power consumption of all the convolution layers in AlexNet \cite{alexnet} using the proposed gather method over the repetitive unicast method. For $8 \times 8$ mesh, we have less than 1\% improvement for all the convolution layers. However, we can see this improvement is better with the increased size of the network. For $16 \times 16$ mesh, the total improvement in all the convolution layers is around 8\%. Similarly, Fig. \ref{fig-power-vgg16} shows the performance improvement of power in VGG-16 \cite{vgg16}. We can see a similar trend, with an increase in the network size the performance improvement is getting better.

\begin{table}[]
	\caption{Convolution layers for AlexNet \cite{alexnet} \& VGG-16 \cite{vgg16}}
	\begin{center}

	\begin{tabular}{|c|c|c|c|}
		\hline
		Model                    & Layers & \begin{tabular}[c]{@{}c@{}}Kernels\\ (C $\times$ \#kernels @ R $\times$ R)\end{tabular} & \begin{tabular}[c]{@{}c@{}}Layer Size\\ (C @ H $\times$ H)\end{tabular} \\ \hline
		\multirow{6}{*}{AlexNet \cite{alexnet}} & Input  &                                                                           & 3@224x224                                                        \\ \cline{2-4} 
		& Conv1  & 3x64@11x11                                                                & 64@55x55                                                         \\ \cline{2-4} 
		& Conv2  & 64x192@5x5                                                                & 192@27x27                                                        \\ \cline{2-4} 
		& Conv3  & 192x384@3x3                                                               & 384@13x13                                                        \\ \cline{2-4} 
		& Conv4  & 384x256@3x3                                                               & 256@13x13                                                        \\ \cline{2-4} 
		& Conv5  & 256x256@3x3                                                               & 256@13x13                                                        \\ \hline
		\multirow{5}{*}{VGG-16 \cite{vgg16}}  & Input  &                                                                           & 3@224x224                                                        \\ \cline{2-4} 
		& Conv1  & 64x64@3x3                                                                 & 64@224x224                                                       \\ \cline{2-4} 
		& Conv2  & 128x128@3x3                                                               & 128@112x112                                                      \\ \cline{2-4} 
		& Conv3  & 256x256@3x3                                                               & 256@56x56                                                        \\ \cline{2-4} 
		& Conv4  & 512x512@3x3                                                               & 512@14x14                                                        \\ \hline
	\end{tabular}

	\end{center}
	\label{table-conv}
\end{table}

\section{Conclusion} \label{conclusion}
In this paper, we proposed to use the gather packet on mesh-based NoCs to support the significant amount of many-to-one traffic existing in convolution and max pooling layers of a CNN workload. Particularly during the convolution layer, continuous streams of input and weights are fed to PEs and the MAC results are sent back from each PE to the output buffer before starting the new round of computation. The simulation results of the proposed method using the traffic traces generated from the convolution layers of AlexNet and VGG-16 are compared against the repetitive unicast method on both $8 \times 8$ and $16 \times 16$ mesh-based NoCs. AlexNet \cite{alexnet} and VGG-16 \cite{vgg16} both show the improvement in the total runtime and power consumption for $8 \times 8$ and $16 \times 16$ mesh-based NoCs employing output stationary systolic array. The proposed gather method can be used to support gather type traffic in other workloads. However, to prevent the time out of $\delta$ when mixed with other traffic a separate VC can be allocated to the gather traffic. In our future work, this method will be tested with other types of dataflow models, workloads, and combined with other routing support methods to accelerate the complete CNN model. 
\bibliography{ref}
\bibliographystyle{IEEEtran}
\end{document}